\documentclass[10pt,twocolumn,letterpaper]{article}

\usepackage[pagenumbers]{cvpr} 

\usepackage[dvipsnames]{xcolor}

\definecolor{cvprblue}{rgb}{0.21,0.49,0.74}
\usepackage[pagebackref,breaklinks,colorlinks,citecolor=cvprblue]{hyperref}

\def\paperID{*****} 
\def\confName{CVPR}
\def\confYear{2024}

\title{Bootstrap Segmentation Foundation Model under Distribution Shift \\via Object-Centric Learning}

\author{{Luyao Tang$^{1}$$^{*}$, Yuxuan Yuan$^{1}$$^{*}$, Chaoqi Chen$^{2}$, Kunze Huang$^{1}$, Xinghao Ding$^{1}$ and Yue Huang$^{1}$$^{\dag}$}\\
{$^1$ School of Informatics, Xiamen University, China}\\
{$^2$ Department of Computer Science, The University of Hong Kong, China}\\
{\tt\small  \{lytang, yuanyuxuan0908\}@stu.xmu.edu.cn, cqchen1994@gmail.com, kzhuang@stu.xmu.edu.cn}\\
{\tt\small  \{dxh, yhuang2010\}@xmu.edu.cn}
}

\begin{document}
\maketitle
\begin{abstract}

Foundation models have made incredible strides in achieving zero-shot or few-shot generalization, leveraging prompt engineering to mimic the problem-solving approach of human intelligence. However, when it comes to some foundation models like Segment Anything, there is still a challenge in performing well on out-of-distribution data, including camouflaged and medical images. Inconsistent prompting strategies during fine-tuning and testing further compound the issue, leading to decreased performance. Drawing inspiration from how human cognition processes new environments, we introduce \textbf{SlotSAM}, a method that reconstructs features from the encoder in a self-supervised manner to create object-centric representations. These representations are then integrated into the foundation model, bolstering its object-level perceptual capabilities while reducing the impact of distribution-related variables. The beauty of SlotSAM lies in its simplicity and adaptability to various tasks, making it a versatile solution that significantly enhances the generalization abilities of foundation models. Through limited parameter fine-tuning in a bootstrap manner, our approach paves the way for improved generalization in novel environments. The code is available at \href{https://github.com/lytang63/SlotSAM}{github.com/lytang63/SlotSAM}.

\end{abstract}
\vspace*{-7mm}
\footnote{$^{*}$First two authors contributed equally.}\footnote{$^{\dag}$Corresponding author.}
\section{Introduction}
\label{sec:intro}

\begin{figure}[t]
  \centering \includegraphics[width=0.95\linewidth]{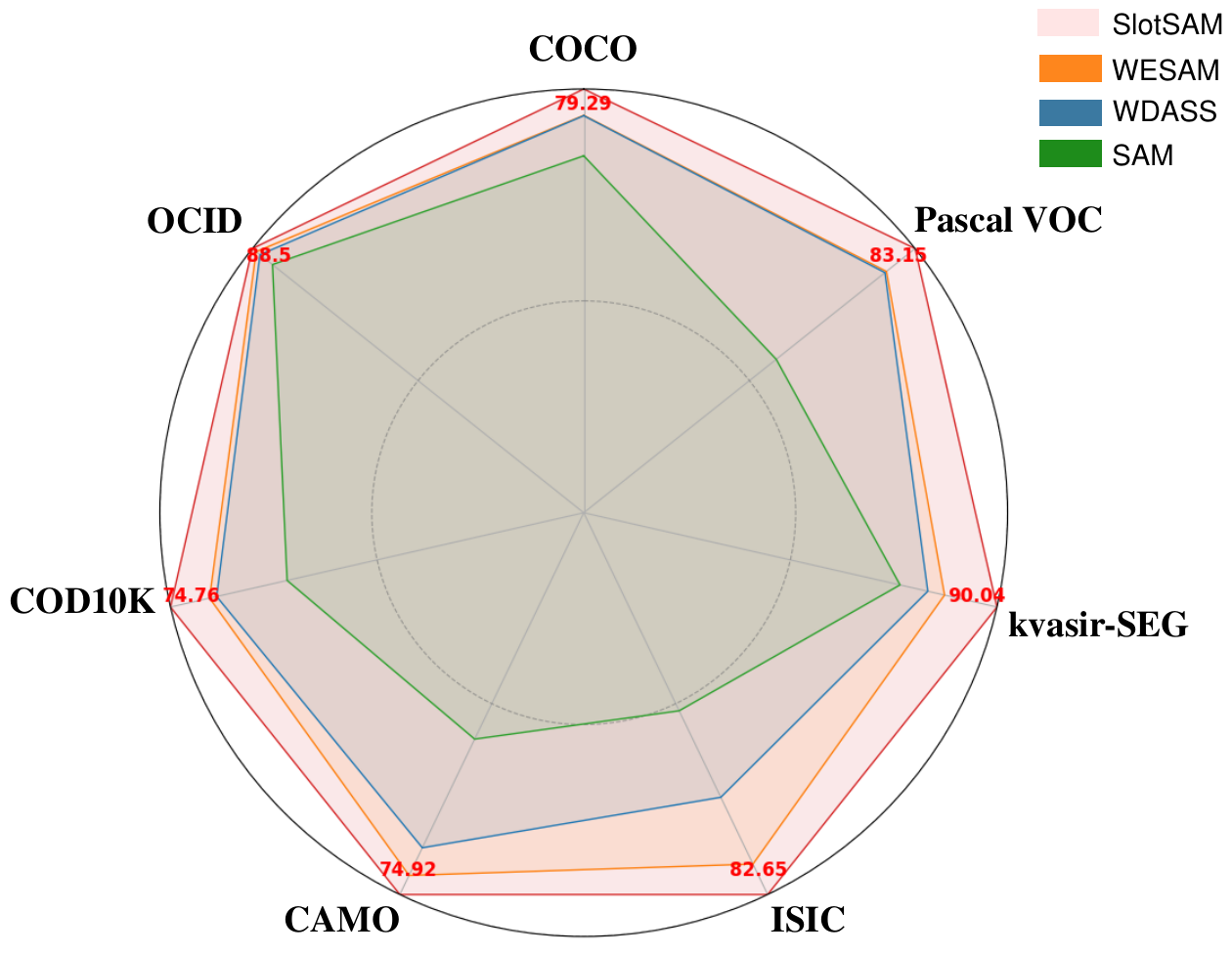}
   \caption{Performance comparison between SAM, WDASS, WESAM and SlotSAM across different downstream tasks under distribution shift and prompt shift.}
   \vspace{-0.4cm}
   \label{fig:lidar}
\end{figure}

\subsection{Existing Studies and Limitations}


The impressive capabilities of foundation models~\cite{kirillov2023segment, ke2024segment, radford2021learning, roziere2023code, touvron2023llama} in zero-shot learning are a significant factor in their growing prominence. Taking the segmentation foundation models as an example, their primary goal is to achieve strong performance in dense predictions on arbitrary images, with the Segment Anything Model (SAM) ~\cite{kirillov2023segment} being a representative work. Despite SAM's claimed robust zero-shot segmentation capabilities, distribution shift in challenging downstream tasks (e.g., medical imaging, camouflaged objects, low-quality images) undermines its advantages.

Enhancing SAM's generalization and robustness on new data is a key focus. Fine-tuning is an intuitive method to adapt SAM to various downstream tasks. This may involve customizing a medical image-specific adapter~\cite{MedSAM} or integrating SAM as an additional supervisory branch in a semi-supervised segmentation framework for improved consistency learning~\cite{zhang2023semisam}. However, these techniques require retraining on datasets with fine-grained annotations, often unavailable in real-world scenarios.

\begin{figure*}[t]
  \centering \includegraphics[width=0.95\linewidth]{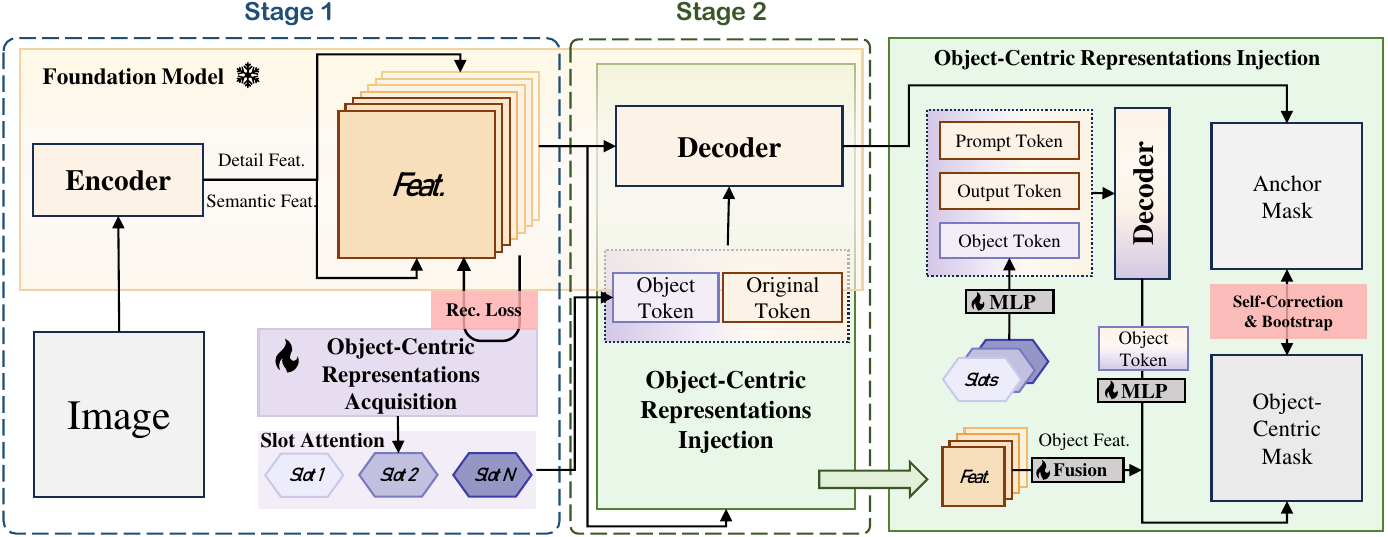}
   \caption{Overview of SlotSAM. Stage 1 is to obtain slots by reconstructing higher-order semantics. Stage 2 is to inject slots into the foundation model by nonlinearly combining them into object token and self-training. The whole process is task-independent.}
   \label{fig:overview}
   \vspace{-0.3cm}
\end{figure*}

Recent research~\cite{li2024asam} has employed Stable Diffusion to enhance a subset of the SA-1B~\cite{kirillov2023segment} dataset by generating adversarial instances, leading to improved performance of the SAM model, which requires unsustainable consumption of resources. Although this strategy constitutes a form of model fine-tuning, its implementation in real-world scenarios remains challenging due to substantial computational resource consumption. WESAM~\cite{zhang2024improving} focuses on adapting the SAM by incorporating a frozen source model as an anchor. Under weak supervision, it utilizes LoRA~\cite{hu2021lora} to fine-tune the model, thereby diminishing reliance on data and computational resources. However, WESAM's enforcement of contrastive learning between different instances within images disrupts the semantic relationships among similar objects and can result in error accumulation.

\subsection{Our Intuition and Insights}

The poor performance of current foundation models in unknown environments can be attributed to two types of real-world shifts. The first is \textbf{distribution shift}~\cite{koh2021wilds, taori2020measuring}, which occurs when the data used for training (source domain) has a different distribution from the data encountered in downstream tasks during the actual application (target domain). The second is \textbf{prompt shift}~\cite{abdul2024align, zhou2018brief}, where downstream tasks provide only coarse weak supervision instead of the fine-grained labels available in the source domain.

To address these challenges, we draw inspiration from the perceptual pipeline of human cognition in unfamiliar environments. We aim to simulate human-like intelligence~\cite{burns2023weak} by abstracting the real world at the object level and injecting this capability into any foundation model. Object-centric learning~\cite{locatello2020object} operates based on causal mechanisms that align with the physical world. By leveraging its combinatorial reasoning properties in scene comprehension, object-centric learning reduces reliance on domain-specific variables and enables more robust handling of out-of-distribution data.

Applying Slot-Attention~\cite{locatello2020object}, the core technology of object-centric learning, to unsupervised RGB pixel reconstruction in foundation models lacks meaningfulness for three reasons. Firstly, the optimization objective of reconstructing the image itself lacks sufficient information for discerning real-world objects, potentially leading to degradation, as shown in~\cref{fig:slots}. Secondly, training foundation models typically involve large-sized images, resulting in unacceptable resource overhead associated with Slot-Attention. Finally, the injection of object-centric representations compatible with foundation models and the enhancement of their object perception capabilities warrant careful consideration.

Considering the aforementioned factors, our objective is to redefine the reconstruction target of Slot-Attention as high-level features with stronger inductive biases. The image encoder of the foundation model effectively extracts high-level semantics for each object within the image, offering a uniform representation of the high-dimensional nature of the real world without being biased by pixel color reconstruction. High-quality object-centric representations, seamlessly integrated with existing tokens in most foundation models, can be considered as object tokens. During the forward process, object tokens can leverage the attention mechanism among tokens to access global image context, geometric region, semantic information, and mask regions. This significantly enhances the foundation model's object perception capabilities with minimal fine-tuning parameters. As the entire process is unsupervised and reinforces the generalizability of the foundation model, relying on its exceptional feature representation, we define it as bootstrapping. Our contributions can be summarized as follows: 

\textit{we introduce a task-agnostic approach for acquiring high-quality object-centric representations, enhancing the generalization capability of foundation models by incorporating object perception capabilities while keeping resource consumption low.}

\begin{table*}
  \centering
     \scalebox{0.575}
 {
    \begin{tabular}{c|c|c|c|c|c|c|c|c|c|c|c|c|c|c|c|c|c|c|c|c|c}
    \hline \multirow{2}{*}{ Method } & \multicolumn{3}{|c|}{ \cellcolor{blue!10}COCO 2017 } & \multicolumn{3}{c}{ \cellcolor{blue!10}Pascal VOC } & \multicolumn{3}{|c|}{ \cellcolor{green!10}kvasir-SEG } & \multicolumn{3}{|c|}{ \cellcolor{green!10}ISIC } &  \multicolumn{3}{|c|}{ \cellcolor{yellow!10}CAMO } & \multicolumn{3}{|c|}{ \cellcolor{yellow!10}COD10K } & \multicolumn{3}{|c}{ \cellcolor{red!10}OCID }\\
    & \cellcolor{gray!15}box & \cellcolor{gray!30}point & \cellcolor{gray!45}poly & \cellcolor{gray!15}box & \cellcolor{gray!30}point & \cellcolor{gray!45}poly & \cellcolor{gray!15}box & \cellcolor{gray!30}point & \cellcolor{gray!45}poly & \cellcolor{gray!15}box & \cellcolor{gray!30}point & \cellcolor{gray!45}poly & \cellcolor{gray!15}box & \cellcolor{gray!30}point & \cellcolor{gray!45}poly & \cellcolor{gray!15}box & \cellcolor{gray!30}point & \cellcolor{gray!45}poly & \cellcolor{gray!15}box & \cellcolor{gray!30}point & \cellcolor{gray!45}poly\\
    \hline 
    SAM~\cite{kirillov2023segment} & 74.29 & 55.06 & 65.64 & 69.21 & 69.21 & 60.79 & 81.59 & 62.30 & 54.03 & 66.74 & 53.42 & 62.82  & 62.72 & 57.43 & 50.85 & 66.32 & 63.61 & 40.04 & 86.35 & 71.41 & 72.81\\
    \hline 
    TENT~\cite{wang2021tent} & 78.21 & 52.99 & 71.51 & 80.24 & 74.97 & 65.03  & 82.47 & 61.84 & 62.97 & 71.76 & 53.46 & 67.12 & 71.24 & 59.59 & 60.29 & 69.36 & 61.94 & 43.36 & 87.77 & 66.61 & 77.53\\
    SHOT~\cite{liang2021source} & 75.18 & 58.46 & 69.26 & 79.80 & 74.26 & 63.38 & 82.30 & 63.76 & 61.34 & 71.99 & 55.99 & 66.86 & 71.61 & 62.78 & 58.72 & 69.09 & 65.25 & 42.38 & 88.06 & 74.39 & 76.25  \\
    TRIBE~\cite{su2024towards} & 77.56 & 49.56 & 70.99 & 78.87 & 69.21 & 65.39 & 85.05 & 73.03 & 64.61 & 72.61 & 50.36 & 67.99 & 66.00 & 61.97 & 60.54 & 67.84 & 63.62 & 42.75 & 86.77 & 67.86 & 76.50 \\
    DePT~\cite{gao2022visual} & 71.00 & 37.35 & 63.27 & 74.09 & 42.99 & 59.94 & 81.91 & 52.06 & 61.55 & 78.43 & 46.79 & 72.75  & 55.44 & 33.07 & 48.63 & 59.32 & 34.06 & 35.51 & 82.00 & 56.52 & 70.92\\
    WDASS~\cite{das2023weakly} & 77.29 & 60.55 & 70.19 & 80.12 & 76.15 & 66.98 & 84.01 & 63.78 & 64.78 & 74.23 & 55.63 & 67.84  & 71.25 & 63.39 & 62.29 & 71.42 & 65.61 & 43.93 & 87.68 & 77.13 & 76.70\\
    WESAM~\cite{zhang2024improving} & 77.32 & 60.5 & 70.77 & 80.27 & 74.15 & 66.72 & 85.47 & 75.23 & 67.40 & 80.01 & 62.12 & 75.36  & 73.42 & 65.55 & 62.90 & 71.93 & 70.55 & 45.87 & 88.09 & 80.14 & 77.41 \\
    \hline
    Ours & $\mathbf{79.29}$ & $\mathbf{60.99}$ & $\mathbf{75.48}$ & $\mathbf{83.15}$ & $\mathbf{77.23}$ & $\mathbf{70.77}$ & $\mathbf{90.04}$ & $\mathbf{81.96}$ & $\mathbf{79.64}$ & $\mathbf{82.65}$ & $\mathbf{66.21}$ & $\mathbf{78.72}$  & $\mathbf{74.92}$ & $\mathbf{68.95}$ & $\mathbf{71.09}$ & $\mathbf{74.76}$ & $\mathbf{72.46}$ & $\mathbf{48.86}$ & $\mathbf{88.50}$ & $\mathbf{81.35}$ & $\mathbf{79.54}$ \\
     improv. & +5.00 & +5.93 & +9.84 & +13.94 & +8.02 & +9.98 & +8.45 & +19.66 & +25.61 & +15.91 & +12.79 & +15.90  & +12.20 & +11.52 & +20.24 & +8.44 & +8.85 & +8.82 & +2.15 & +9.94 & +6.73 \\
    \hline 
    Supervised & 81.50 & 69.77 & 73.39 & 81.23 & 76.98 & 71.32 & 85.89 & 77.54 & 81.64 & 81.62 & 79.81 & 80.26 & 79.17 & 77.01 & 67.12 & 78.06 & 78.44 & 64.90 & 91.24 & 89.22 & 79.23 \\
    \hline
    \end{tabular}
}
  \caption{Comparison with SOTAs on \colorbox{blue!10}{\color{black}natural}, \colorbox{green!10}{\color{black}medical}, \colorbox{yellow!10}{\color{black}camouflaged object}, and \colorbox{red!10}{\color{black}robotic} image datasets using \colorbox{gray!15}{\color{black}bounding box}, \colorbox{gray!30}{\color{black}sparse points}, and \colorbox{gray!45}{\color{black}coarse segmentation mask} prompts. The results of the baselines in the table are from~\cite{zhang2024improving}.}
  \label{tab:tabs_1}
\end{table*}

\section{Methodology}
\label{sec:method}

\subsection{Preliminaries}

Our original intention is to provide a way to inject object-centric representation perception capabilities into foundation models in a general sense. Therefore, the training process of the foundation models is not the focus of attention, so we do not differentiate between the optimization objectives of the original foundation models or the fine-tuned foundation models, modeling their loss functions as $\mathcal{L}_{\text{base}}$.

We chose the SAM as a representative foundation model for our research. SAM consists of three main components: the image encoder $\mathbf{z}=f(\mathbf{x} ; \Theta)$, the prompt encoder $\mathbf{e}=g(\mathbf{p} ; \Omega)$, and the mask decoder $h(\mathbf{z}, \mathbf{e} ; \Phi)$. Inspired by~\cite{zhang2024improving}, we used the universal optimization method as $\mathcal{L}_{\text{base}}$ for SlotSAM. We maintain three encoder networks, where for each input $\mathbf{x}$, we obtain $\mathbf{x}_s$ and $\mathbf{x}_w$ through strong augmentation and weak augmentation, respectively. $\mathbf{x}_w$ is then processed by the anchor model $f(\mathbf{x}_w ; \Theta^a)$ and the student model $f(\mathbf{x}_w ; \Theta^s)$ to obtain $\mathcal{M}^a$ and $\mathcal{M}^s$, while the teacher model $f(\mathbf{x}_s ; \Theta^t)$ processes $\mathbf{x}_w$ to obtain $\mathcal{M}^t$. $\mathcal{M}$ represents the predicted mask. A generic teacher-student self-training loss can be defined as

\begin{equation}
\mathcal{L}_{\text{base}}=\mathcal{L}^{\text {dice}}\left(\mathcal{M}^{s/t}, \mathcal{M}^a\right) + \mathcal{L}^{\text {focal}}\left(\mathcal{M}^s, \mathcal{M}^t\right)  .
\end{equation}


\subsection{Object-Centric Representation Acquisition}

Simply reconstructing RGB pixels allows Slot-Attention to achieve some effectiveness on synthetic datasets, but in the real world, RGB supervision signals are insufficient to represent objects and environments, making them prone to degradation, as shown in~\cref{fig:slots}. Inspired by~\cite{seitzer2023bridging}, object-centric representation requires a more well-trained semantic encoder, and fortunately, the encoder of the foundation model can provide rich semantic details.

The underlying logic of Slot-Attention is to reconstruct features through self-supervision, compressing high-dimensional, semantically rich, and unstructured object features into low-dimensional structured information in a bottleneck-like manner. Slots act as the bottleneck, retaining object-centric representation. Therefore, given the output feature $\mathbf{z} \in \mathbb{R}^{N \times D_z}$ from the encoder $f(\mathbf{x} ; \Theta)$, and initializing a set of slots $\mathbf{s} \sim \mathcal{N}(\mathbf{s} ; \boldsymbol{\mu}, \boldsymbol{\sigma}) \in \mathbb{R}^{K \times D_s}$, $K$ is the number of slots, $D_z$ and $D_s$ represent the dimension of output feature and slot. We project them to dimensionds by a linear transformation $\mathcal{K}_\beta$ for slots and $\mathcal{Q}_\gamma$, $\mathcal{V}_\phi$ for $\mathbf{z}$, and the Slot-Attention is trained as $\operatorname{update}(\boldsymbol{A}, \mathbf{v})=\boldsymbol{A}^T \mathbf{v}$, 
where $\operatorname{update}(\boldsymbol{A}, \mathbf{v})=\boldsymbol{A}^T \mathbf{v}, A_{i j}=\frac{\operatorname{attn}(\mathbf{q}, \mathbf{k})_{i j}}{\sum_{l=1}^K \operatorname{attn}(\mathbf{q}, \mathbf{k})_{l j}}, \operatorname{attn}(\mathbf{q}, \mathbf{k})=\frac{e^{M_{i j}}}{\sum_{l=1}^N e^{M_{i l}}}, \boldsymbol{M}=\frac{\mathbf{k q}^T}{\sqrt{D_s}}$. The $\mathbf{q}=\mathcal{Q}_\gamma(\mathbf{z}) \in \mathbb{R}^{K \times D_s}, \mathbf{k}=\mathcal{K}_\beta(\mathbf{z}) \in \mathbb{R}^{N \times D_s}$, and $\mathbf{v}=\mathcal{V}_\phi(\mathbf{z}) \in \mathbb{R}^{N \times D_s}$ denote the query, key and value vectors respectively. The query is a function of the slots. After optimizing $T$ iterations using the Gated Recurrent Unit~\cite{chung2014empirical, dey2017gate} (GRU), the slots are passed through a slot-decoder to output the reconstructed feature $\mathbf{\hat{z}}$, minimizing the self-supervised reconstruction loss:

\begin{equation}
\mathcal{L}_{\text {rec}}=\|\mathbf{\hat{z}}-\mathbf{z}\|^2, \quad \mathbf{\hat{z}}=\text { slot-decoder }(\mathbf{s}) .
\end{equation}

$\mathbf{\hat{z}}$ is the weighted sum of each slots. Since each slot should be associated with a different object, each slot should be able to attend to specific spatial regions. Following the approach in~\cite{seitzer2023bridging}, we employ an efficient MLP as a spatial broadcast decoder~\cite{watters2019spatial}. Each slot is broadcasted to several patches with the addition of positional encoding. The tokens for each slot are processed individually by the MLP, and after channel division, we obtain the reconstructed feature $\mathbf{\hat{z}}_k$ and the activation region ${\alpha}_k$. The weighted feature representation $\mathbf{\hat{z}}$ for all slots after reconstruction is obtained by

\begin{equation}
\mathbf{\hat{z}}=\sum_{k=1}^K \mathbf{\hat{z}}_k \odot \boldsymbol{m}_k, \quad \boldsymbol{m}_k=\underset{k}{\operatorname{softmax}}({\alpha}_k) .
\end{equation}

\subsection{Object-Centric Representation Injection}

In the decoder of SAM, the predicted mask is obtained by performing element-wise multiplication between the Output Token $\mathcal{T}_{out} \in \mathbb{R}^{N_{out} \times D_{out}}$ and the mask feature. The accuracy of the mask is strongly correlated with the amount of information provided by the tokens. Therefore, as shown in~\cref{fig:overview}, we innovatively design the object-centric representation stored in the slots to be the Object Token. This design is fully compatible with the original decoder architecture, and thanks to the attention mechanism, the Object Token can exchange information with other tokens. The Object Token can access the global image's contextual information and geometric details. Furthermore, the existing $\mathcal{T}_{out}$ can acquire more discriminative features related to objects, such as positional information and topological associations.

For each input $\mathbf{x}$, there is a corresponding set of slots $\mathbf{s}$. To avoid disrupting the optimization preference established by the decoder for existing tokens, $\mathbf{s}\in \mathbb{R}^{K \times D_s}$ is fed into an MLP for nonlinear combination to obtain the Object Token $\mathcal{T}_{obj} \in \mathbb{R}^{1 \times D_s}$, where $D_s = D_{out}$. In each attention layer, the Object Token performs self-attention calculations with other tokens and shares the same feed-forward layers to ensure consistent optimization direction of model.

As the $\mathcal{T}_{obj}$ contains more deep semantic features and fewer detailed features, introducing local boundary details helps avoid boundary blurring for objects. Inspired by~\cite{ke2024segment}, We extract the detail features from the first attention block of the encoder. After applying transposed convolution, we add the detail feature with the semantic features to obtain fused object features $\mathbf{z}^{obj}$. Then, similar operations are performed, where the $\mathcal{T}_{obj}$ is multiplied by the $\mathbf{z}^{obj}$ to obtain the Object-Centric Mask $\mathcal{M}^o$.

Self-training networks may suffer from the problem of error accumulation due to incorrect predictions. Therefore, in the early stages of training, we fix the parameters of the anchor model (with $\mathbf{x}_w$ as the input). The trained model is referred to as the object-centric model (with $\mathbf{x}_s$ as the input). We use a simplified loss function in the style of $\mathcal{L}_{\text {base}}$ to train the MLP and Fusion modules, in order to prevent significant bias in knowledge transfer:

\begin{equation}
    \mathcal{L}^{\text {dice}}\left(\mathcal{M}^{o}, \mathcal{M}^a\right) + \mathcal{L}^{\text {bce}}\left(\mathcal{M}^o, \mathcal{M}^a\right) .
\end{equation}

In the later stages of training, we employ a bootstrap strategy. At the end of epochs where the model has improved its mIoU on the validation set, we directly copy the parameters of the object-centric model to the anchor model. Through this iterative process, we gradually complete the bootstrap of the foundation model.

\section{Experiments}

\textbf{Quantitative analysis:} As shown in~\cref{tab:tabs_1} and~\cref{fig:lidar}, we evaluated SlotSAM on seven datasets and three prompt modalities. Astonishingly, SlotSAM outperforms existing methods by a large margin. SlotSAM narrows the gap on natural images with fully supervised fine-tuning and even surpassed fine-grained masking under point or poly prompts supervision. On medical images, our mIoU on the kvasir-SEG dataset exceeded 90\% and remained performant even under poly prompts, surpassing WESAM~\cite{zhang2024improving} by 18.16\%. On the most challenging dataset of camouflaged objects, we achieved an average improvement of over 3\%.

\begin{figure}[h]
  \centering \includegraphics[width=0.95\linewidth]{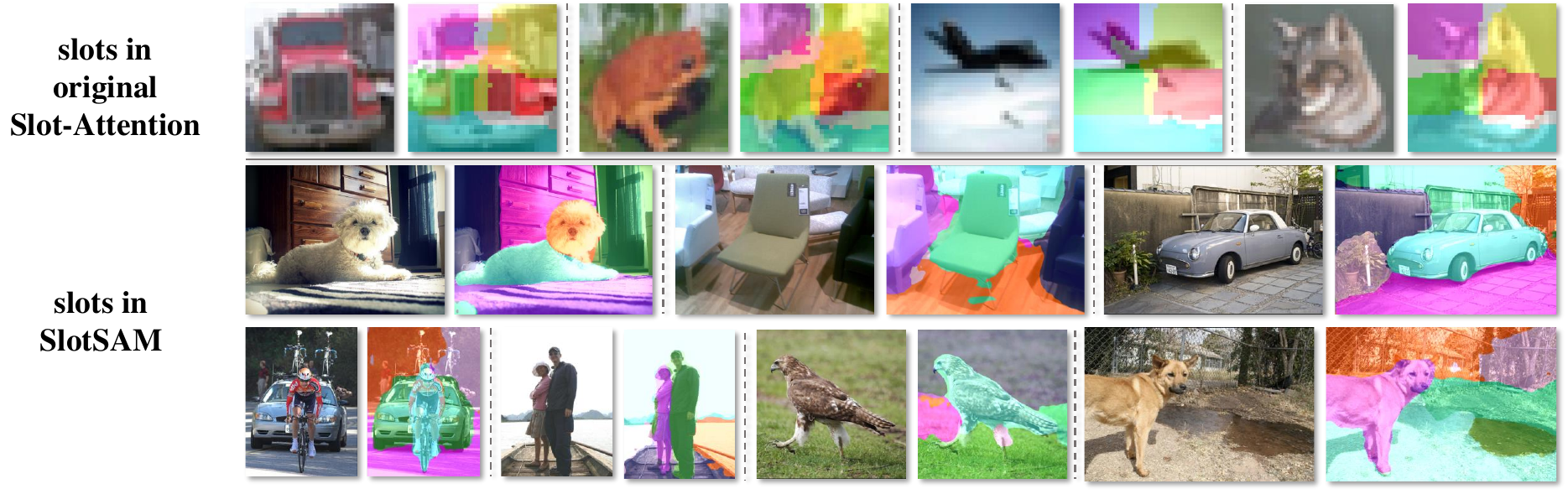}
   \caption{Comparison of the quality of the slots.}
   \label{fig:slots}
   \vspace{-0.2cm}
\end{figure}

\begin{figure}[h]
  \centering \includegraphics[width=0.95\linewidth]{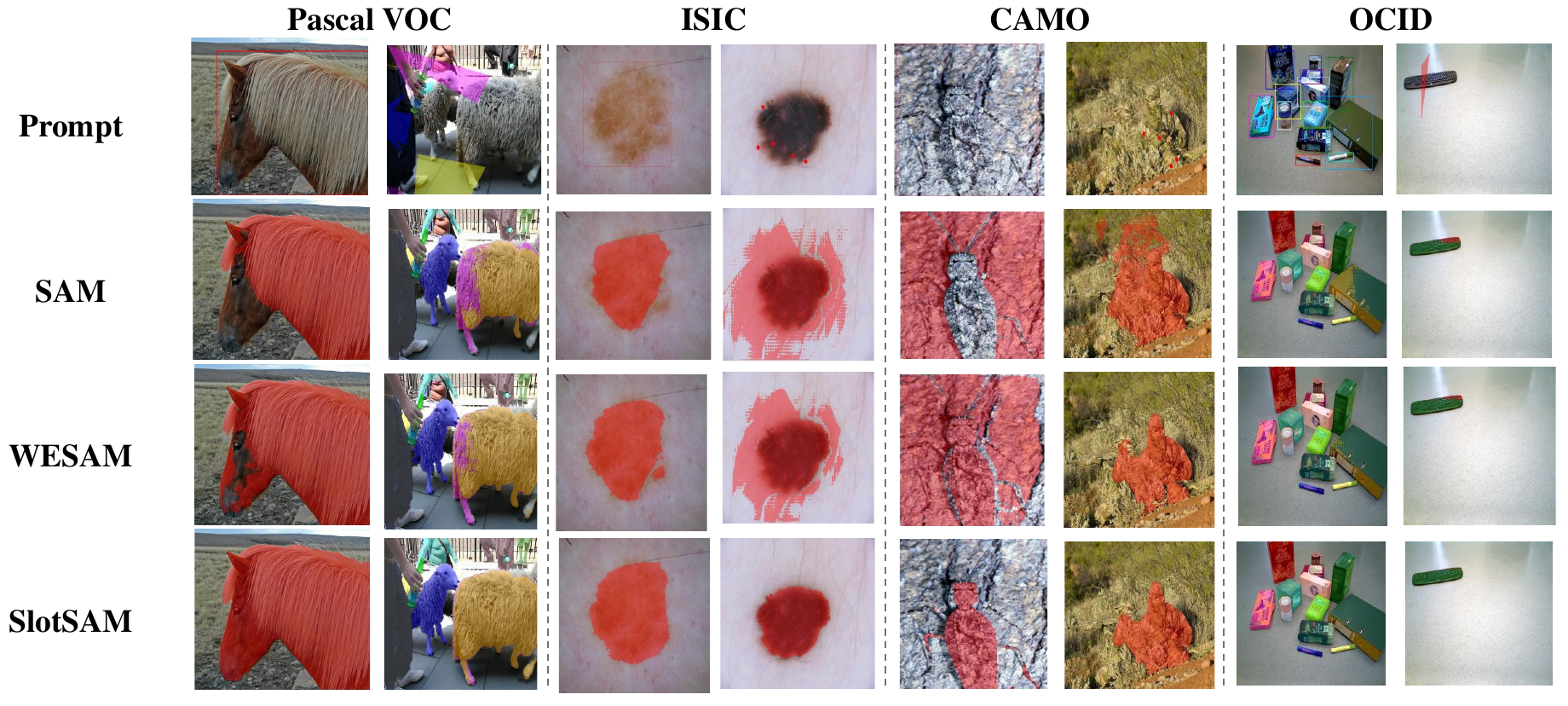}
   \caption{Comparison of the fineness of the predicted masks.}
   \vspace{-0.2cm}
   \label{fig:masks}
\end{figure}

\textbf{Qualitative analysis:} \cref{fig:masks} compares the masks predicted by SlotSAM and existing SOTA methods. SlotSAM provides the most detailed predictions in regions with smaller pixel area occupancy, such as the junction of a horse's hair and face, indicating its ability to capture the semantic correlations within the object and incorporate the acceptable boundaries into the object representation. SlotSAM provides higher semantic distinctiveness in ambiguous boundary areas, such as camouflaged objects and similar backgrounds, avoiding semantic confusion. This demonstrates that the distinctiveness among slots is transferred to the foundation model, enabling it to distinguish between different objects easily. \cref{fig:slots} illustrates that SlotSAM can obtain non-degraded object-centric representations.

\section{Conclusion and Discussion }

We research the generalization of foundation models under distribution shift. A universal, concise approach is proposed to inject object-centric representation awareness into the foundation model, enabling it to focus on causal factors related to objects and achieve generalization. Building upon the framework of SAM, we propose SlotSAM, which can self-train in a bootstrapping manner without accessing the fine-grained label. The proposed method has undergone extensive evaluations in multiple datasets that cover four types of downstream tasks. The experimental results have demonstrated that SlotSAM has taken an important step toward robustly handling distribution shift and prompt shift.

In this work, we primarily focus on acquiring and injecting object-centric representations. Future research could explore differences in the distribution of slots, potentially enabling models to have more detailed awareness and it would be meaningful to utilize slots for controlled generative models.
{
    \small
    \bibliographystyle{ieeenat_fullname}
    \bibliography{main}
}


\end{document}